\documentclass[conference]{IEEEtran}
\IEEEoverridecommandlockouts
\usepackage{amsmath,amssymb,amsfonts}
\usepackage{algorithm}
\usepackage{algorithmic,bm}
\usepackage{graphicx}
\usepackage{textcomp}
\usepackage{xcolor}
\usepackage[style=ieee]{biblatex} 
\usepackage{ifthen}

\makeatletter
\newcounter{IEEE@bibentries}
\renewcommand\IEEEtriggeratref[1]{%
  \renewbibmacro{finentry}{%
    \stepcounter{IEEE@bibentries}%
    \ifthenelse{\equal{\value{IEEE@bibentries}}{#1}}
    {\finentry\@IEEEtriggercmd}
    {\finentry}%
  }%
}
\makeatother

\def\q{{\boldsymbol q}}
\def\k{{\boldsymbol k}}
\def\v{{\boldsymbol v}}

\def\x{{\boldsymbol x}}
\def\W{{\boldsymbol W}}
\def\R{{\boldsymbol R}}

\def\K{{\boldsymbol K}}
\def\V{{\boldsymbol V}}
\def\N{{\boldsymbol N}}

\def\BibTeX{{\rm B\kern-.05em{\sc i\kern-.025em b}\kern-.08em
    T\kern-.1667em\lower.7ex\hbox{E}\kern-.125emX}}
\begin{document}
\title{Accurate Block Quantization in LLMs with Outliers}
\author{\IEEEauthorblockN{Nikita Trukhanov}
\IEEEauthorblockA{\textit{d-Matrix}\\
Santa Clara, CA, USA\\
ntrukhanov@d-matrix.ai}
\and
\IEEEauthorblockN{Ilya Soloveychik}
\IEEEauthorblockA{\textit{d-Matrix}\\
Santa Clara, CA, USA\\
ilyas@d-matrix.ai}
}
\maketitle
\begin{abstract}
The demand for inference on extremely large scale LLMs has seen enormous growth in the recent months. It made evident the colossal shortage of dedicated hardware capable of efficient and fast processing of the involved compute and memory movement. The problem is aggravated by the exploding raise in the lengths of the sequences being processed, since those require efficient on-chip storage of the KV-cache of size proportional to the sequence length. To make the required compute feasible and fit the involved data into available memory, numerous quantization techniques have been proposed that allow accurate quantization for both weights and activations. One of the main recent breakthroughs in this direction was introduction of the family of Block Floating Point (BFP) formats characterized by a block of mantissas with a shared scale factor. These enable memory- power-, and compute- efficient hardware support of the tensor operations and provide extremely good quantization accuracy. The main issues preventing widespread application of block formats is caused by the presence of outliers in weights and activations since those affect the accuracy of the other values in the same block. In this paper, we focus on the most critical problem of limited KV-cache storage. We propose a novel approach enabling usage of low precision BFP formats without compromising the resulting model accuracy. We exploit the common channel-wise patterns exhibited by the outliers to rearrange them in such a way, that their quantization quality is significantly improved. The methodology yields 2x savings in the memory footprint without significant degradation of the model's accuracy. Importantly, the rearrangement of channels happens at the compile time and thus has no impact on the inference latency.
\end{abstract}


\section{Introduction}
Pretrained Large Language Models (LLMs) have become enormously popular in the recent years \cite{zhang2022opt, touvron2023llama, openai2024gpt4, jiang2024mixtral,team2024gemma}. Such popularity has mostly been gained due to the extremely high quality of the text generated by the state-of-the-art models. However, such improvements often come at the cost of increased model sizes which makes training of these large models and using them for inference highly challenging in terms of storage capacity, memory transfer, and compute. The architecture of the modern LLMs is typically based on the decoder part of a transformer \cite{vaswani2017attention}. While the LLM training process can fully exploit parallelization across the input tokens, the inference must be performed sequentially. The generation process produces one token on every pass over the network given the prompt and all previously generated tokens. The core building block of the transformer architecture -- the attention mechanism -- requires computation of the so called keys $\K$ and values $\V$ representing the information stored in the entire sequence on every generative step. When the sequences become too large, repetitive computations of the $\K$ and $\V$ matrices become prohibitively resource greedy. To avoid those redundant operations, one could exploit the fact that the keys and values of the already appended tokens never change and can therefore be cached on chip.

Caching $\K$ and $\V$ matrices is extremely helpful if the on-chip storage allows it. However, the ever growing demand for generation of longer sequence dwarfs any amount of on-chip storage \cite{hooper2024kvquant, ding2024longrope}. Hence, every possible technique must be exploited to reduce the memory footprint of the cached tensors. The most promising approach consists in efficient quantization of keys and values. To this end such algorithms as GPTQ \cite{frantar2023gptq}, SmoothQuant \cite{xiao2023smoothquant}, and many others have been proposed. For example, the GPTQ technique prescribes successive quantization of the weight columns in such a way that the rounding of every next column carefully takes into account the accumulated error of the previously quantized columns. The error is calculated on a small representative batch of data. In contrast, SmoothQuant is targeted to better quantization of activations. The authors notice that in the activations they were observing, a few channels had consistently higher values on various tokens. They introduced per-channel scaling factors to carry the dynamic range of activations over into weights. This way they transferred part of quantization burden from harder-to-quantize activations to easier-to-quantize weights. In all quantization approaches, the goal is always to enable a low-bit, e.g. 4 bits per element, storage for the tensors, with a common scaling vector, and, in some cases, bias vectors.

Further refinements of the algorithmic and software solutions have only limited impact on the overall efficiency if not supported by hardware. Most of the modern LLM models are designed and run on Graphics Processing Unit (GPUs) which exploit floating-point arithmetic \cite{wang2019benchmarking, srinivas2021bottleneck}. As mentioned earlier, the computational load required by modern transformers has reached such enormous volumes that traditional GPUs cannot fully meet the growing demand, pushing both accelerators and high performance GPUs towards narrow arithmetic. As a consequence, unmatched research efforts have been applied by the engineering community to replace narrow floating-point with even denser fixed-point representations \cite{zadeh2020gobo, zafrir2019q8bert, shen2020q, zhang2020ternarybert}. Despite the excellent gains in both speed and computational density achieved by fixed-point arithmetic, training using it or even half-precision floating-point arithmetic has not provided clear evidence in its favor due to the limited dynamic range inherent in such formats \cite{micikevicius2017mixed}.

Block Floating Point (BFP) numerical formats have received renewed interest recently for LLM inference applications due to their combination of wide dynamic range, numerical accuracy, and efficient hardware implementation of inner products using simple integer arithmetic \cite{darvish2020pushing, lyubomirsky2022clock, soloveychik2022block, mxfp2023mx}. BFP formats are characterized by a block of mantissas with a shared scale factor. The simplest implementation has the scale factor as a power of two, the so-called exponent, in which case the inner product between two blocks involves multiplying the integer mantissas and adding the two block exponents. The industry has thus far mainly exploited BFP12 (with $4$-bit element mantissas and $8$ bit shared exponent) and BFP16 (with $8$-bit element mantissas and $8$ bit shared exponent), both of which can be used with different block sizes usually ranging from $16$ to $128$ elements \cite{lyubomirsky2022clock, soloveychik2022block, darvish2020pushing, mxfp2023mx}. Alternative formats, using low-bit floating point elements, with a wider range common exponent, are also considered \cite{rouhani2023microscaling}. 

One of the main numerical issues faced by the ML engineers dealing with LLMs both from theoretical and practical perspectives is the sporadic emergence of so-called outliers in weights and activations of the modern large-scale transformers \cite{xiao2023smoothquant, hooper2024kvquant}. Existence of outliers becomes especially challenging when it comes to block formats, since presence of even a single element with an extremely large magnitude in a block can completely ruin the quantization accuracy of all the other elements in that same block.

Below, we address this problem. We demonstrate how the advantages of the BFP quantization can be maintained when weights or activations contain numerous outliers. The key observation behind our approach consists in the fact that the inner product is invariant to synchronized reshuffling of the tensors being multiplied. For instance, if we focus on the $\q\K^\top$ product we can easily see that permuting the channels of the keys and queries simultaneously in exactly same manner has no impact on the product. Since the keys $\K$ are outputs of the linear layer with weights $\W_\k$, the values of each channel are determined by the corresponding row of $\W_\k$ and the relevant inputs. Now we can simply rearrange the channels (rows) of $\W_\k$ in such a way that will make its block quantization very accurate. This permutation must be compensated by the reshuffling of $\W_\q$ which does not affect the accuracy since $\q$ is anyway computed in high precision and is not stored in the cache. Note that the reordering of the channels in $\W_\k$ and $\W_\q$ happens at the compile time. It requires no calibration data and has no effect on the inference latency.

The rest of the paper is organized as follows. In section \ref{inference} we describe the setup in more detail and define the block formats. Section \ref{algo} features our novel $\K$-sort algorithm that allows accurate low-precision BFP-quantization of $\K$ cache containing outliers. Supporting empirical data is provided in Section \ref{experiment}. We summarize our findings in Section \ref{conclusion}.

\section{Inference in LLMs} \label{inference}
In this paper, we focus on the problem of inference in LLMs. The sizes of the up-to-date models have become so large and the amount of compute involved became so enormous that efficient processing requires dedicated hardware and specialized algorithms.

\subsection{KV-cache}
Inference on modern transformers essentially means sequential generation of tokens one by one given the initial prompt. After every pass through the model's stack of decoders, the newly generated token is appended to the growing sequence and the process repeats with the updated context. The very nature of the attention mechanism requires calculation of the keys and values for the entire sequence generated up until current iteration. This leads to a lot of duplicated compute since every head inside every decoder block will repeatedly calculate the entire $\K$ and $\V$ tensors for all the tokes over and over again. In order to avoid the expensive recomputations, one usually stores the $\K$ and $\V$ values of the already generated tokens in the cache memory. Then on every following iteration the $\q$, $\k$, and $\v$ values only for the currently processed token are computed while the rest of $\K$ and $\V$ matrices are retrieved from the cache. With growing sequence length, we can easily get out of on-chip memory. In this article, we suggest a computationally efficient way to decrease the memory footprint of the $\K$ cache. Importantly, our approach leads to only very minor accuracy loss even when $\K$ contains numerous outliers.

\subsection{Block Floating Point Formats}
The unprecedented and ever growing amount of compute and storage required by the modern LLMs has lead to the development of numerous new data formats and novel directions and techniques involving quantization of weights and activations. New data formats are announced every few months both by the computer science community training the models \cite{ma2024era, peng2023fp8} and by the manufacturers of hardware \cite{darvish2020pushing, rouhani2023microscaling, micikevicius2022fp8, mxfp2023mx}. Different techniques are proposed separately for storage and for compute \cite{hooper2024kvquant, peng2023fp8, ma2024era}.

In this work, we focus on an extremely promising Block Floating Point family of formats that has become very popular in the recent months \cite{darvish2020pushing, lyubomirsky2022clock, soloveychik2022block, mxfp2023mx}. The idea is based on the observation that quite often the elements of involved tensors have comparable amplitudes and thus can share the same or close exponent value when written in floating-point notation. As a consequence, we can store entire blocks of elements using shared exponent and individual integer mantissas. Numerous companies design there hardware specifically to support this family of formats \cite{darvish2020pushing, lyubomirsky2022clock, soloveychik2022block}. The main advantage enjoyed by the chips designed to support BFP formats consists in very significant reduction of required storage and effectively integer matrix multiplication, see \cite{lyubomirsky2022clock, soloveychik2022block} for more details. This further leads to a huge reduction in consumed power and energy.

More specifically, a Block Floating Point format is characterized by the block size $n \in \mathbb{N}$, mantissa precision $p \in \mathbb{N}$, and precision of its exponent $b$. All the elements $\{M_i\}_{i=1}^n$ of a block are stored as integers in the range of $[-(2^{p-1}-1), 2^{p-1}-1]$, and their values are computed as
\begin{equation}
\{2^e\cdot M_1,\dots,2^e\cdot M_n\},
\end{equation}
where $e$ is a $b$-bit integer.

Blocks formats are extremely efficient for matrix operations, since dot product using this family of formats effective turns into integer matrix multiplication and simple addition of the corresponding block exponents \cite{soloveychik2022block, darvish2020pushing, mxfp2023mx}. The typical values of $p$ are usually $4$ and $8$ bits per elements and the corresponding formats read as BFP12 and BFP16. The block sizes often range from $16$ to $128$ \cite{soloveychik2022block, darvish2020pushing, mxfp2023mx}. Casting an array into BFP format requires computation of the block exponent based on the largest absolute value element inside the block, scaling the block elements based on this exponent and then rounding the resulting values to the closest integers in the mantissa range $[-(2^{p-1}-1), 2^{p-1}-1]$.

\subsection{Sorting Channels of $\W_\k$}
To efficiently store matrix $\K$ in the cache and compute $\q\K^T$ faster, we propose to quantize the former into a low-precision block format. The definition of the BFP format, says that the quantization range of a block is determined by its largest (in absolute value) element. If some blocks contain outliers, their overall quantization accuracy will be poor because the smallest elements might be rounded to zero. Next we show how to resolve this problem.

The natural approach would be to sort the elements of the tensor by their absolute values before quantization. In that case, each block will only contain elements of comparable magnitudes: there will be blocks with larger elements and blocks with smaller elements, but we will avoid the undesirable scenario of having numerous blocks containing mixtures of elements of wide dynamic range. However, we must note that sorting tensors on the fly would be prohibitively expensive. Also, if we need to keep the sorting order to restore the original one for every token for every attention layer, it would outweigh any memory savings. Therefore, the brute-force sorting of elements will not work and we need a finer approach.

\begin{figure*}[htbp]
\centerline{\includegraphics[width=\textwidth]{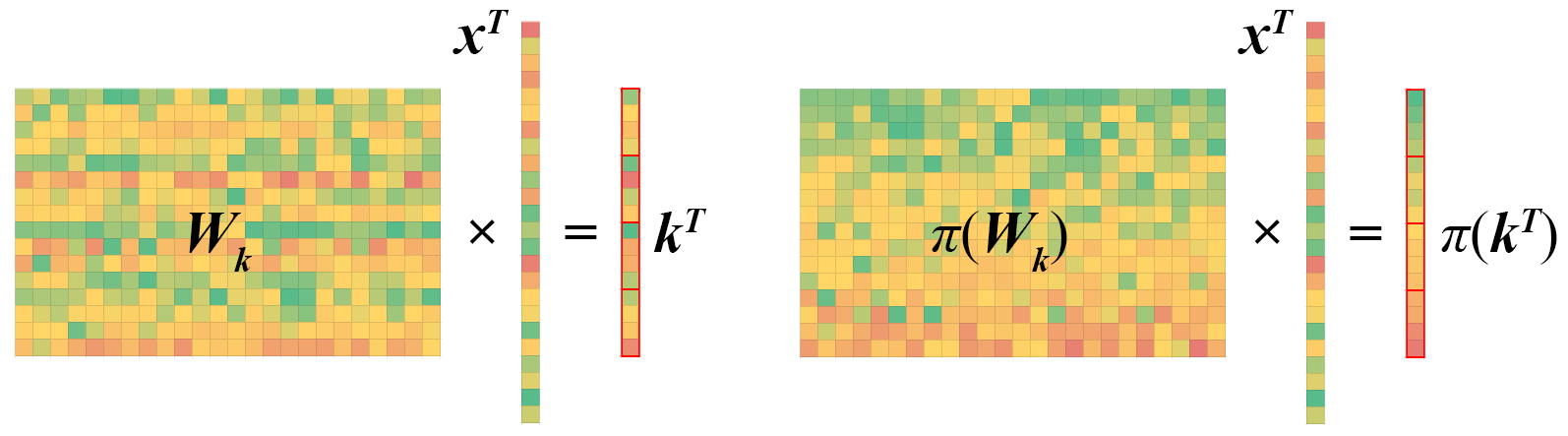}}
\caption{Left: original $\W_\k$ and $\k$. Right: rows of $\W_\k$ have been sorted by their Euclidean norms to yield $\pi(\W_\k)$ and the resulting $\pi(\k^\top)$; colors reflect the absolute values of the elements, from lower (green) to larger (red); BFP quantization of $\pi(\k^\top)$ is more accurate than that of $\k^\top$ since the entries of the former ending up in same blocks are closer in their absolute values.}
\label{wxt}
\end{figure*}

As noted in \cite{hooper2024kvquant}, the keys tend to exhibit certain outlier patterns. Namely, the outliers often concentrate in particular channels, which are quite consistent both across tokens in input sequences, and across different input sequences. Such behavior is usually caused by higher norms of the corresponding rows of $\W_\k$ projection matrices. We can, therefore, easily sort the channels of $\W_\k$ by their Euclidean norms before the inference starts. To compensate for this reshuffling, we also rearrange $\W_\q$ in the same order. Lets denote the corresponding permutation by $\pi$, then due to the linearity of inner product we can say that
\begin{equation}
\W_q^\top\cdot\W_\k = [\pi(\W_\q)]^\top\cdot\pi(\W_\k).
\end{equation}
As a consequence, we have
\begin{multline}
\q\cdot\k^\top = \x\W_\q^\top\cdot\W_\k\x^\top =  \x[\pi(\W_\q)]^\top\cdot\pi(\W_\k)\x^\top \\ =[\pi(\q^\top)]^\top\cdot\pi(\k^\top).
\end{multline}
Now that we have applied permutation $\pi$ to the static weights $\W_\k$ and $\W_\q$ at the compile time, the permuted vectors $\pi(\q^\top)$ and $\pi(\k^\top)$ will be automatically computed during inference and their product will be exactly equal to the original $\q\cdot\k^\top$. The idea is illustrated by Figure \ref{wxt}. The colors of the heat-map reflect the absolute values of the elements, from lower (green) to larger (red). It is important to note that we do not store the queries in the cache and they can therefore be cast to a higher precision format. To enable application of this technique to any transformer, we need to show how it works when rotary embeddings are applied to keys and queries.

\subsection{Rotary Embeddings}
Many modern LLMs use rotary positional embeddings (RoPE) \cite{su2024roformer} to encode information about the order of tokens in the input sequence. Rotary embeddings are linear transformations applied to keys and queries defined as 
\begin{equation*}
\R_{\Theta,m}^{d_h} =
	\begin{pmatrix}
		\cos{m\theta_1}& -\sin{m\theta_1}&\cdots&0\\
		\sin{m\theta_1}&\cos{m\theta_1}&\cdots&0 \\
		\vdots&\ddots&\ddots&\vdots\\
		0&\cdots&\cos{m\theta_{d_h/2}}& -\sin{m\theta_{d_h/2}}\\
		0&\cdots&\sin{m\theta_{d_h/2}}&\cos{m\theta_{d_h/2}}
	\end{pmatrix},
\end{equation*}
where $m$ is the token index, and $\theta_i,\, i \in 1..d_h/2$ are predefined constants with $d_h$ - the head dimension. Matrices $\R_{\Theta,m}^{d_h}$ define block-diagonal orthogonal transformations rotating 2-dimensional linear subspaces spanned by every consecutive pair of standard basis vectors. Due to the evident sparsity of $\R_{\Theta,m}^{d_h}$, the multiplication is more efficiently implemented as the following linear transformation of the the input vector $\x$,
\begin{multline}
	\R^{d_h}_{\Theta, m}\x = 
	\begin{pmatrix}
		x_1\\
		x_2\\
		x_3\\
		x_4\\
		\vdots\\
		x_{d_h-1}\\
		x_{d_h}
	\end{pmatrix}
	\otimes
	\begin{pmatrix}
		\cos{m\theta_1} \\
		\cos{m\theta_1} \\
		\cos{m\theta_2} \\
		\cos{m\theta_2} \\
		\vdots \\
		\cos{m\theta_{d_h/2}} \\
		\cos{m\theta_{d_h/2}} 
	\end{pmatrix}
	\\
        + 
	\begin{pmatrix}
		-x_2\\
		x_1\\
		-x_4\\
		x_3\\
		\vdots\\
		-x_{d_h}\\
		x_{d_h-1}
	\end{pmatrix}
	\otimes
	\begin{pmatrix}
		\sin{m\theta_1}\\
		\sin{m\theta_1}\\
		\sin{m\theta_2}\\
		\sin{m\theta_2}\\
		\vdots\\
		\sin{m\theta_{d_h/2}}\\
		\sin{m\theta_{d_h/2}}
	\end{pmatrix},
    \label{RoFreq}
\end{multline}
where $\otimes$ is the element-wise product. Next we provide a general version of our sorting algorithm that works well when rotary embeddings are used.

\section{K-sort Algorithm}
\label{algo}
The main idea of our $\K$-sort algorithm is to sort the rows $\W_\k$ according to their norms (increasing or decreasing order). We called the required permutation of row indices $\pi$. This same permutation is then used to reshuffle the rows of $\W_\q$ and both reorderings happen at the compile time. When rotary embeddings are used, we need to add two natural auxiliary steps to make sure the locations of the elements of $\R^{d_h}_{\Theta, m}$ are changed correctly. For that we need to correctly reorder the vector of frequencies $\Theta = [\theta_1, \theta_1, \theta_2, \theta_2, \dots, \theta_{d_h/2}, \theta_{d_h/2}]$ and keep track of the sine-signs. The original order of the sine-sign channel indices is given by equation (\ref{RoFreq}), $\bm{i} = [2,1,4,3,\dots,d_h, d_h-1]$, the corresponding signs read as $\bm{s} = [-1,1,-1,1,\dots,-1,1]$\footnote{We note that in the open-source implementations of models such as LLama2 \cite{touvron2023llama}, the order of channels is different, with the frequency vector being ${[\theta_1, \theta_2, \dots, \theta_{d_h/2}, \theta_1, \theta_2, \dots, \theta_{d_h/2}]}$. It requires reordering of $\bm{i}$ and $\bm{s}$ but does not affect our algorithm, hence we quote the original RoPE ordering.}. We emphasize that since the reordering permutation $\pi$ is known at the compile time, all the necessary permutations of the frequencies and signs needed for correct application of RoPE can be done then as well - this does not delay the inference. 

\begin{algorithm}
\centering
    \caption{$\K$-sort algorithm for a head}
    \small
    \label{alg1}
    \begin{algorithmic}
    \STATE 1: $\N_i \gets ||\W_\k[i,:] ||, \; \forall i \in \{1,\dots,d_h\}$
    \STATE 2: $\pi \gets  \text{argsort}\left(\N \right)$
    \STATE 3: $\W_\k \gets \pi\left(\W_\k \right)$
    \STATE 4: $\W_\q \gets \pi\left(\W_\q \right)$
    \STATE // \textit{the last step apply to models with rotary embeddings}
    \STATE 5: $\Theta \gets \pi\left(\Theta\right), \; \bm{i} \gets \pi(\bm{i}), \; \bm{s} \gets \pi(\bm{s})$
    \end{algorithmic}
\end{algorithm}

In practice, we propose to use $\K$-sort algorithm with BFP12 quantization of the keys. More specifically, matrix $\K$ inside each head is stored in block format with $4$ bits of precision per element and a shared exponent for a block of $32, 64$ or $128$ elements. This allows 2x compression of the cache versus the more common $8$-bit storage without much loss of accuracy (see Section \ref{experiment} for empirical evidence). On every token generation pass, once the keys are retrieved from the cache the correctly ordered rotary embedddings are applied to them and the product with $\pi(\q)$ is computed. Since on the generative stage the $\pi(\q)$ tensor is very small in size and we dont need to store it, it can be easily computed in the higher precision BFP16 format with $8$-bits per element mantissa without any significant effect on the performance.

While present work concentrates on the keys $\K$, similar technique can be applied to the values $\V$. Moreover, since there are no rotary embeddings involved there, permuting the rows of matrix $\W_\v$ simultaneously with the columns of the following projection layer is much easier to implement and that will not affect the output. Such approach will improve the quantization quality of values $\V$ stored in the cache without any run-time overhead. For the lack of space, we postpone the details for further publications.

\section{Experiments}
\label{experiment}
Numerous recent publications have reported the issues of outliers in K-cache and their significant impact on the accuracy and storage requirements \cite{xiao2023smoothquant, hooper2024kvquant, dettmers2022llmint8}. For the lack space, in this short contribution we focus on one of such popular LLMs, Llama2-7B-hf model \cite{touvron2023llama}. As shown in \cite{hooper2024kvquant}, this network and its many relatives and variations exhibit the K-outliers phenomenon very clearly. In this section, we demonstrate the advantages of our $\K$-sort algorithm on this model. Importantly, this network exploits the aforementioned rotary embeddings to encode positional information, which even better utilizes the flexibility of our technique. Llama-7B is a mid-size model but as shown below it can already benefit tremendously from the application of $\K$-sort. This implies that the gain on larger models will be even more remarkable.

The experiments were carried out using the default Hugging Face checkpoint without extra fine-tuning. The baseline perplexity of the model with FP16 weights on wikitext-2 \cite{merity2016pointer} dataset is $9.4881$. We quantized all the keys $\K$ and queries $\q$ into the BFP format with rounding to the nearest. Two formats we used in this setup are BFP12 for the keys and BFP16 for the queries. Both come with $8$-bit shared exponents per block and $4$ or $8$ bits per integer mantissa of the block elements, respectively. Importantly, on the auto-regressive stage of generation, the query tensors are usually small, thus their high-precision quantization makes no impact on performance. In addition, $\q$-s are not stored in the cache so their compression is not required. For fair comparison, the rest of the operations were performed exactly as in the baseline model - in FP16 format. 

Table \ref{tab1} demonstrates the obtained results. As a sanity check, we see that for the block size of $128$, rearranging the channels does not help. This is because the head dimension of Llama2-7B is exactly $128$ and sorting the rows cant help. However, when the block size decreases, the gain provided by $\K$-sort becomes evident. Already for the block size of $64$ we can see a significant improvement in the accuracy after reordering the rows of $\W_\k$.

\begin{table}[htbp]
\caption{LLama2-7B perplexity on wikitext-2}
\begin{center}
\begin{tabular}{|c|c|c|c|}
\hline
\multicolumn{2}{|c|}{\textbf{format}}&\multicolumn{2}{|c|}{\textbf{algorithm}} \\
\hline 
\textbf{Q} & \textbf{K}& \textbf{original}& \textbf{$\K$-sorted} \\
\hline
FP16&FP16&9.4881&9.4881\\
\hline
BFP16\_128&BFP12\_128&10.0861&10.0861\\
\hline
BFP16\_64&BFP12\_64&9.9999&9.6061\\
\hline
BFP16\_32&BFP12\_32&9.8300&9.5196\\
\hline
\end{tabular}
\label{tab1}
\end{center}
\end{table}

\section{Conclusion}
\label{conclusion}
In this paper, we demonstrate that simple reshuffling of the static weights in popular LLMs can make their quantization quality much better. Specifically, we advocate for the use of Block Floating Point formats and show that BFP12 format with $4$-bit mantissa storage and compute without any elaborate quantization enables extremely accurate inference. Our $\K$-sort algorithm together with BFP12 storage allows for 2x reduction of the memory footprint of the $\K$-cache and therefore allows generation of much longer sequences on the same hardware.
\IEEEtriggeratref{25}
\printbibliography

@article{vaswani2017attention,
  title={Attention is all you need},
  author={Vaswani, A. and Shazeer, N. and Parmar, N. and Uszkoreit, J. and Jones, L. and Gomez, A. N. and Kaiser, L. and Polosukhin, I.},
  journal={Advances in Neural Information Processing Systems},
  volume={30},
  year={2017}
}

@article{zafrir2019q8bert,
  title={Q8{BERT}: Quantized 8bit {BERT}},
  author={Zafrir, O. and Boudoukh, G. and Izsak, P. and Wasserblat, M.},
  journal={Workshop on Energy Efficient Machine Learning and Cognitive Computing-NeurIPS Edition},
  pages={36--39},
  year={2019},
  organization={IEEE}
}

@article{shen2020q,
  title={Q-BERT: {h}essian based ultra low precision quantization of {BERT}},
  author={Shen, S. and Dong, Z. and Ye, J. and Ma, L. and Yao, Z. and Gholami, A. and Mahoney, M. W. and Keutzer, K.},
  journal={Proceedings of the AAAI Conference on Artificial Intelligence},
  volume={34},
  number={05},
  pages={8815--8821},
  year={2020}
}

@article{zhang2020ternarybert,
  title={Ternary{BERT}: {d}istillation-aware ultra-low bit {BERT}},
  author={Zhang, W. and Hou, L. and Yin, Y. and Shang, L. and Chen, X. and Jiang, X. and Liu, Q.},
  journal={arXiv preprint arXiv:2009.12812},
  year={2020}
}

@article{lyubomirsky2022clock,
  title={Block Floating Point ({BFP}) for Efficient Deep Neural Net Inference},
  author={Lyubomirsky, I. and Wang, X.},
  journal={IEEE P3109 Working Group, June 6},
  year={2022}
}

@article{micikevicius2017mixed,
  title={Mixed precision training},
  author={Micikevicius, P. and Narang, S. and Alben, J. and Diamos, G. and Elsen, E. and Garcia, D. and Ginsburg, B. and Houston, M. and Kuchaiev, O. and Venkatesh, G. and others},
  journal={arXiv preprint arXiv:1710.03740},
  year={2017}
}

@article{soloveychik2022block,
      title={Block Format Error Bounds and Optimal Block Size Selection}, 
      author={Soloveychik, I. and Lyubomirsky, I. and Wang, X. and Bhoja, S.},
      year={2022},
      journal={arXiv preprint arXiv:2210.05470}
}

@article{su2024roformer,
  title={{R}o{F}ormer: {e}nhanced transformer with rotary position embedding},
  author={Su, Jianlin and Ahmed, Murtadha and Lu, Yu and Pan, Shengfeng and Bo, Wen and Liu, Yunfeng},
  journal={Neurocomputing},
  volume={568},
  pages={127063},
  year={2024},
  publisher={Elsevier}
}

@article{rouhani2023microscaling,
      title={Microscaling Data Formats for Deep Learning}, 
      author={Bita Darvish Rouhani and Ritchie Zhao and Ankit More and Mathew Hall and Alireza Khodamoradi and Summer Deng and Dhruv Choudhary and Marius Cornea and Eric Dellinger and Kristof Denolf and Stosic Dusan and Venmugil Elango and Maximilian Golub and Alexander Heinecke and Phil James-Roxby and Dharmesh Jani and Gaurav Kolhe and Martin Langhammer and Ada Li and Levi Melnick and Maral Mesmakhosroshahi and Andres Rodriguez and Michael Schulte and Rasoul Shafipour and Lei Shao and Michael Siu and Pradeep Dubey and Paulius Micikevicius and Maxim Naumov and Colin Verrilli and Ralph Wittig and Doug Burger and Eric Chung},
      year={2023},
      journal={arXiv preprint arXiv:2310.10537}
}

@article{hooper2024kvquant,
      title={{K}{V}Quant: {t}owards 10 Million Context Length {L}{L}{M} Inference with {K}{V} Cache Quantization}, 
      author={Coleman Hooper and Sehoon Kim and Hiva Mohammadzadeh and Michael W. Mahoney and Yakun Sophia Shao and Kurt Keutzer and Amir Gholami},
      year={2024},
      journal={arXiv preprint arXiv:2401.18079}
}

@article{xiao2023smoothquant,
      title={{S}mooth{Q}uant: {a}ccurate and Efficient Post-Training Quantization for Large Language Models}, 
      author={Guangxuan Xiao and Ji Lin and Mickael Seznec and Hao Wu and Julien Demouth and Song Han},
      year={2023},
      journal={arXiv preprint arXiv:2211.10438}
}

@article{touvron2023llama,
      title={Llama 2: {o}pen Foundation and Fine-Tuned Chat Models}, 
      author={Hugo Touvron and Louis Martin and Kevin Stone and Peter Albert and Amjad Almahairi and Yasmine Babaei and Nikolay Bashlykov and Soumya Batra and Prajjwal Bhargava and Shruti Bhosale and Dan Bikel and Lukas Blecher and Cristian Canton Ferrer and Moya Chen and Guillem Cucurull and David Esiobu and Jude Fernandes and Jeremy Fu and Wenyin Fu and Brian Fuller and Cynthia Gao and Vedanuj Goswami and Naman Goyal and Anthony Hartshorn and Saghar Hosseini and Rui Hou and Hakan Inan and Marcin Kardas and Viktor Kerkez and Madian Khabsa and Isabel Kloumann and Artem Korenev and Punit Singh Koura and Marie-Anne Lachaux and Thibaut Lavril and Jenya Lee and Diana Liskovich and Yinghai Lu and Yuning Mao and Xavier Martinet and Todor Mihaylov and Pushkar Mishra and Igor Molybog and Yixin Nie and Andrew Poulton and Jeremy Reizenstein and Rashi Rungta and Kalyan Saladi and Alan Schelten and Ruan Silva and Eric Michael Smith and Ranjan Subramanian and Xiaoqing Ellen Tan and Binh Tang and Ross Taylor and Adina Williams and Jian Xiang Kuan and Puxin Xu and Zheng Yan and Iliyan Zarov and Yuchen Zhang and Angela Fan and Melanie Kambadur and Sharan Narang and Aurelien Rodriguez and Robert Stojnic and Sergey Edunov and Thomas Scialom},
      year={2023},
      journal={arXiv preprint arXiv:2307.09288}
}

@article{frantar2023gptq,
      title={{G}{P}{T}{Q}: {a}ccurate Post-Training Quantization for Generative Pre-trained Transformers}, 
      author={Elias Frantar and Saleh Ashkboos and Torsten Hoefler and Dan Alistarh},
      year={2023},
      journal={arXiv preprint arXiv:2210.17323}
}

@article{merity2016pointer,
  title={Pointer sentinel mixture models},
  author={Merity, Stephen and Xiong, Caiming and Bradbury, James and Socher, Richard},
  journal={arXiv preprint arXiv:1609.07843},
  year={2016}
}

@article{darvish2020pushing,
  title={Pushing the limits of narrow precision inferencing at cloud scale with microsoft floating point},
  author={Darvish Rouhani, Bita and Lo, Daniel and Zhao, Ritchie and Liu, Ming and Fowers, Jeremy and Ovtcharov, Kalin and Vinogradsky, Anna and Massengill, Sarah and Yang, Lita and Bittner, Ray and others},
  journal={Advances in neural information processing systems},
  volume={33},
  pages={10271--10281},
  year={2020}
}

@article{wang2019benchmarking,
  title={Benchmarking {TPU}, {GPU}, and {CPU} platforms for deep learning},
  author={Wang, Y. E. and Wei, G.-Y. and Brooks, D.},
  journal={arXiv preprint arXiv:1907.10701},
  year={2019}
}

@article{srinivas2021bottleneck,
  title={Bottleneck transformers for visual recognition},
  author={Srinivas, A. and Lin, T.-Y. and Parmar, N. and Shlens, J. and Abbeel, P. and Vaswani, A.},
  journal={IEEE/CVF Conference on Computer Vision and Pattern Recognition},
  pages={16519--16529},
  year={2021}
}

@article{zadeh2020gobo,
  title={{GOBO}: {q}uantizing attention-based {NLP} models for low latency and energy efficient inference},
  author={Zadeh, A. H. and Edo, I. and Awad, O. M. and Moshovos, A.},
  journal={IEEE/ACM International Symposium on Microarchitecture},
  pages={811--824},
  year={2020},
  organization={IEEE}
}

@article{openai2024gpt4,
      title={{G}{P}{T}-4 Technical Report}, 
      author={OpenAI},
      year={2024},
      journal={arXiv preprint arXiv:2303.08774}
}

@article{jiang2024mixtral,
      title={Mixtral of Experts}, 
      author={Albert Q. Jiang and Alexandre Sablayrolles and Antoine Roux and Arthur Mensch and Blanche Savary and Chris Bamford and Devendra Singh Chaplot and Diego de las Casas and Emma Bou Hanna and Florian Bressand and Gianna Lengyel and Guillaume Bour and Guillaume Lample and Lélio Renard Lavaud and Lucile Saulnier and Marie-Anne Lachaux and Pierre Stock and Sandeep Subramanian and Sophia Yang and Szymon Antoniak and Teven Le Scao and Théophile Gervet and Thibaut Lavril and Thomas Wang and Timothée Lacroix and William El Sayed},
      year={2024},
      journal={arXiv preprint arXiv:2401.04088}
}

@article{team2024gemma,
  title={Gemma: {o}pen Models Based on {G}emini Research and Technology},
  author={Team, Gemma and Mesnard, Thomas and Hardin, Cassidy and Dadashi, Robert and Bhupatiraju, Surya and Pathak, Shreya and Sifre, Laurent and Rivi{\`e}re, Morgane and Kale, Mihir Sanjay and Love, Juliette and others},
  journal={arXiv preprint arXiv:2403.08295},
  year={2024}
}

@article{zhang2022opt,
  title={{O}{P}{T}: {o}pen pre-trained transformer language models},
  author={Zhang, Susan and Roller, Stephen and Goyal, Naman and Artetxe, Mikel and Chen, Moya and Chen, Shuohui and Dewan, Christopher and Diab, Mona and Li, Xian and Lin, Xi Victoria and others},
  journal={arXiv preprint arXiv:2205.01068},
  year={2022}
}

@article{dettmers2022llmint8,
      title={{L}{L}{M}.int8(): 8-bit Matrix Multiplication for Transformers at Scale}, 
      author={Tim Dettmers and Mike Lewis and Younes Belkada and Luke Zettlemoyer},
      year={2022},
      journal={arXiv preprint arXiv:2208.07339}
}

@article{ma2024era,
  title={The Era of 1-bit {LLM}s: all large language lodels are in 1.58 bits},
  author={Ma, Shuming and Wang, Hongyu and Ma, Lingxiao and Wang, Lei and Wang, Wenhui and Huang, Shaohan and Dong, Li and Wang, Ruiping and Xue, Jilong and Wei, Furu},
  journal={arXiv preprint arXiv:2402.17764},
  year={2024}
}

@article{peng2023fp8,
  title={{FP}8-lm: Training {FP}8 large language models},
  author={Peng, Houwen and Wu, Kan and Wei, Yixuan and Zhao, Guoshuai and Yang, Yuxiang and Liu, Ze and Xiong, Yifan and Yang, Ziyue and Ni, Bolin and Hu, Jingcheng and others},
  journal={arXiv preprint arXiv:2310.18313},
  year={2023}
}

@article{micikevicius2022fp8,
  title={{FP}8 formats for deep learning},
  author={Micikevicius, Paulius and Stosic, Dusan and Burgess, Neil and Cornea, Marius and Dubey, Pradeep and Grisenthwaite, Richard and Ha, Sangwon and Heinecke, Alexander and Judd, Patrick and Kamalu, John and others},
  journal={arXiv preprint arXiv:2209.05433},
  year={2022}
}

@article{mxfp2023mx,
  title={{MX} Pytorch Emulation Library},
  author={Microsoft},
  journal={https://github.com/microsoft/microxcaling},
  year={2023}
}

@article{ding2024longrope,
  title={Long{R}o{PE}: Extending {LLM} Context Window Beyond 2 Million Tokens},
  author={Ding, Yiran and Zhang, Li Lyna and Zhang, Chengruidong and Xu, Yuanyuan and Shang, Ning and Xu, Jiahang and Yang, Fan and Yang, Mao},
  journal={arXiv preprint arXiv:2402.13753},
  year={2024}
}
\end{document}